\pdfoutput=1

\documentclass[11pt]{article}

\usepackage[]{EMNLP2022}

\usepackage{times}
\usepackage{latexsym}

\usepackage{fancyhdr}

\fancypagestyle{alim}{\fancyhf{}\fancyfoot[C]{\emph{\small Published as a conference paper at EMNLP 2022 (Findings)
}}}
\usepackage[utf8]{inputenc}

\usepackage{microtype}

\usepackage{inconsolata}

\usepackage{courier}  
\usepackage{helvet}  
\usepackage{tabularx}
\usepackage{url}
\usepackage{comment}
\usepackage{multirow}
\usepackage{amsmath} 
\usepackage{graphicx}
\usepackage{xspace}
\usepackage{threeparttable}
\usepackage{stmaryrd}
\usepackage{color}
\usepackage{stmaryrd}
\usepackage{xspace}
\usepackage{threeparttable}

\newcommand{\ignore}[1]{}
\usepackage{amsmath}
\usepackage{amssymb}
\usepackage{mathabx}
\usepackage{booktabs}
\usepackage[ruled,linesnumbered]{algorithm2e}
\usepackage{soul}

\title{Improving Generalization of  Pre-trained Language Models via Stochastic Weight Averaging}

\author{Peng Lu$^{1,3}$, Ivan Kobyzev$^1$, Mehdi Rezagholizadeh$^1$, Ahmad Rashid$^{1, 2}$, \\ \textbf{Ali Ghodsi$^2$,  Philippe Langlais$^3$}  \\
 $^1$Huawei Noah’s Ark Lab\\
$^2$ Department of Statistics and Actuarial Science, University of Waterloo \\
$^3$ RALI/DIRO, Universit\'e de Montr\'eal, Canada\\
{\{peng.lu1,ivan.kobyzev,mehdi.rezagholizadeh,ahmad.rashid\}@huawei.com} \\
{ali.ghodsi@uwaterloo.ca, felipe@iro.umontreal.ca}}

\begin{document}
\maketitle
\pagestyle{fancy}
\fancyhf{}
\thispagestyle{alim}
\begin{abstract}



Knowledge Distillation (KD) is a commonly used technique for improving the generalization of compact Pre-trained Language Models (PLMs) on downstream tasks. However, such methods impose the additional burden of training a separate teacher model for every new dataset.
Alternatively, one may directly work on the improvement of the optimization procedure of the compact model towards better generalization. Recent works observe that the flatness of the local minimum correlates well with better generalization.
In this work, we adapt Stochastic Weight Averaging (SWA), a method encouraging convergence to a flatter minimum, to fine-tuning PLMs. We conduct extensive experiments on various NLP tasks (text classification, question answering, and generation) and different model architectures and demonstrate that our adaptation improves the generalization without extra computation cost. Moreover, we observe that this simple optimization technique is able to outperform the state-of-the-art KD methods for compact models.






\end{abstract}
\section{Introduction}
Large Pre-trained Language Models (PLMs) like BERT \citep{bert} and GPT \citep{gpt3} have achieved state-of-the-art (SOTA) performance on a number of NLP applications. A large model pre-trained on a massive text corpus learns general-purpose features of the language and can be easily adapted to different downstream tasks by fine-tuning with task-specific objectives~\citep{bert, roberta, scale_law}.  
Multiple papers~\citep{gpt3, pangu, 1000layers} demonstrated that the improvement in pre-training can be transferred favorably to most downstream tasks with little or no data, which leads to considerable efforts to scale up PLMs and increase the size of the pre-training dataset.
However, a recent work~\citep{up_down_gap} empirically shows that as the upstream accuracy increases, the performance of downstream tasks gradually saturates. 
Furthermore, although PLMs perform well in a task-agnostic few-shot setting, it is still desirable to fine-tune them to achieve SOTA performance, especially for models with limited capacity. This encourages more efforts to improve the generalization of PLMs through additional techniques like Knowledge Distillation or better optimization.


Knowledge Distillation~\citep{kd} is an effective method to improve the generalization of compact PLMs. One can use a larger model (a teacher) to fine-tune a smaller model (a student) by making a student learn the teacher's output. This technique has helped in achieving SOTA performance on multiple Natural Language Understanding (NLU) problems for compact models~\citep{mobile_bert, tiny_bert, distilbert}. However, KD is not an efficient method as it requires training a separate teacher model for every new task. 
Alternatively, some works focus on boosting the performance of the optimization algorithm directly. Several directions have been explored in the literature: adding noise to the gradient descent process \citep{Kleinberg2018AnAV, orvieto22noise}, applying continuation optimization \cite{mollifying}, steering optimization towards flat optima \citep{sam, lp_filter}, averaging checkpoints~\citep{swa, soup}. 
Encouraged by the success of Stochastic Weight Averaging (SWA)~\citep{swa} for computer vision and Sharpness-Aware Minimization (SAM) for NLP~\citep{sam_lm}, we
investigate whether SWA can find better,  generalizable minima for fine-tuning PLMs while maintaining training efficiency. We conduct extensive experiments for PLM fine-tuning including seven NLU tasks, two question answering tasks, and one summarization task with different model architectures. 
Our contributions are as follows: We adapt SWA to NLP and demonstrate that our adaptation can be applied to fine-tuning PLMs without extra computational cost on both classification and generation tasks. Furthermore, we outline several practical recipes that are different from the ones in computer vision including using a high constant learning rate in SWA (instead of a cyclical one) and averaging over steps instead of epochs.

\section{Related Work}
\paragraph{KD for PLMs} 

Knowledge Distillation (KD) is a commonly used technique for improving the generalization of a compact PLM~\citep{mobile_bert, tiny_bert, distilbert}. KD can be thought of as an instance-specific label regularization for the training process~\citep{tfkd, beta_ls}. However, KD is not an efficient technique because it adds the burden of fine-tuning an extra teacher model for every new task.
Moreover, a recent work~\citep{tf_lm} shows that without additional techniques and resources like data augmentation, re-weighting or intermediate layer distillation~\citep{uni_kd, mate_kd, rw_kd, pkd} pure label regularization cannot outperform base fine-tuning for PLMs. 



\paragraph{Geometry of the Loss Landscape}
Loss landscape analysis is a popular method for investigating the model convergence problem~\citep{loss_landscape_nn, bert_lossland, deep_ensembles}. It has been widely studied that neural networks tend to generalize better when they converge to flatter local minima~\citep{flat_nips_94, sharp_minima, vision_transformer}.
Some works inspect the curvature and flatness of the loss surface via Hessian spectrum analysis~\citep{hessian, empirical_hessian} and show the positive correlation between local sharpness and the generalization gap. 
Such observations have inspired research on improving generalization from a geometric perspective~\citep{sam, asam, lp_filter}. 
Another recently proposed method, Stochastic Weight Averaging (SWA)~\citep{swa}, also improves generalization and is easy to implement without significant computational overhead. The original paper explains the better generalization by empirically showing the flatness of the  minimum discovered by SWA. However, all the analysis was done on only computer vision tasks and models. 
To the best of our knowledge, our paper is the first work adapting SWA for fine-tuning PLMs.
\paragraph{Flatness for PLMs}
A recent work~\citep{sam_lm} applies Sharpness-Aware Minimization (SAM)~\citep{sam} to PLM fine-tuning on several NLP tasks. It shows that the flatness-aware optimizer can substantially improve the generalization of language models, but it takes  $1.25\times$ computational cost. Another work~\cite{swa_sam} investigates the relationship between two flat-minima optimizers, SAM and SWA. However, it shows that without corresponding adaptation for pre-trained language models, SWA tends to underperform SAM or even regular fine-tuning on NLP benchmarks.

\begin{algorithm}[!tb]\small
\caption{Stochastic Weight Averaging}\label{alg:swa}
\SetKwData{}{}
\SetKwFunction{Mod}{Mod}  \SetKwFunction{Lr}{Lr}
\SetKwInOut{Input}{input}\SetKwInOut{Output}{output}
\Input{Scheduler \Lr(i), interval length $K$, number of iterations $N$, weights $W$}
\Output{$W_{\text{SWA}}$}
$W \xleftarrow[]{} \hat{W}$ \;
$W_{\text{SWA}} \xleftarrow[]{} W$\;
\For{$i\leftarrow 1$ \KwTo $N$}
{
$\eta$ = \Lr{i}\ \tcp*{step size}
$W \xleftarrow[]{} W - \eta\nabla \mathcal{L}_i(W)$\;
\If {\Mod{i, K} $= 0$} 
{$n_{\text{model}} \xleftarrow[]{} \frac{i}{K}$\ \tcp*{ \# of models}
$W_{\text{SWA}} \xleftarrow[]{} \frac{1}{n_{\text{model}} +1}(W_{\text{SWA}} \times n_{\text{model}} + W)$;}
}
\Return {$W_{\text{SWA}}$}
\end{algorithm}

\section{Revisting SWA}

\begin{table*}[!h]

    \begin{center}
    \begin{small}
     \caption{Performance  on GLUE test sets. For reference we give several knowledge distillation techniques for comparison.}
     \label{tab:glue_test}
    \begin{tabular}{l|cccccccc}
    \toprule
  \textbf{Method}   & \textbf{CoLA} & \textbf{MRPC} & \textbf{RTE}             & \textbf{SST-2} & \textbf{QNLI} & \textbf{QQP} & \textbf{MNLI} & \textbf{Avg.} \\\midrule
\textbf{DistilRoBERTa}        & 53.3          & 87.6          & 72.2                    & 93.3          & 90.4          & 88.9         & 83.5          & 81.3    \\
\textbf{+KD}         & \textbf{54.3}          & 86          & 74.1                     & 93.1          & 91.1         & {89.5}         & 83.6          & 81.7     \\
\textbf{+TAKD} & 53.2 &86.7 & \textbf{74.2} &93.2 & 91 & 89.4 &83.8 &81.6\\
\textbf{+AKD} & {54}            & 88            & 73.7                     & 93.6          & 90.8          & \textbf{89.7}         & 83.8          & 81.9     \\\midrule
\textbf{+SAM}  & 47.5          & 88.8          & 73.6                     & \textbf{94.1}          & 89.6          & 89.1         & 83.8          & 80.9     \\

\textbf{+SWA}         & 53.7          & \textbf{89.5}          & 73.6 & {93.7}          & \textbf{91.4}          & 89.3         & \textbf{84.3}          & \textbf{82.2}  \\\bottomrule 
\end{tabular}
    
    \end{small}
    \end{center}
\end{table*}

We start by briefly reminding the Stochastic Weight Averaging~\citep{swa}, a method which implements efficient weight ensembling during model optimization.
First, a neural network is trained with SGD till it converges to a region close to a local minimum of the training loss landscape. Then, in the second stage, SWA averages the checkpoints along the SGD trajectory using a designated learning rate schedule for better exploration of the parameter space. The original paper considers Cyclical and High Constant learning rate schedules (CHC). During the update, SWA collects model parameters for every $K$ step and averages them for the model ensemble. \citet{cha2021swad} empirically confirms
that SWA can avoid convergence to a sharp local minimum. The training process is shown in Algorithm~\ref{alg:swa}.

\paragraph{CHC learning rate schedule} In every cycle, the learning rate anneals from $\eta_{\text{max}}$ to $\eta_{\text{min}}$. The learning rate for each training iteration $i$ is given by:
\begin{align}
    t_i & = \mod{(i-1, K}) + 1, \nonumber \\
    \eta_i &= (1-\frac{t_i}{K})\eta_{\text{max}} + \frac{t_i}{K} \eta_{\text{min}}, \nonumber
\end{align}
where $K$ is the cycle length and $\eta_{\text{max}} >\eta_{\text{min}}$ are the maximum and minimum learning rate, respectively. When $\eta_{\text{max}} =\eta_{\text{min}}$, the learning rate becomes constant.
\paragraph{SWA for PLMs} 
We accommodate the SWA method for PLM fine-tuning. We observe that several modifications are required compared to SWA for computer vision. 
First, we noticed that a high constant learning rate performes better than a cyclical learning rate. Then, since the fine-tuning needs much less epochs than training from scratch, we average over steps instead of epochs as in the original SWA for computer vision. Besides that, for most of the NLU tasks we average a larger number of components and start averaging (stage 2 of SWA) earlier: after $50 \% $  of the training instead of $75 \%$  in the original method. The latter could be because the PLM is closer to the local minimum than a randomly initialized model and hence allows more exploration.


\section{Experiments}
In this section, we introduce the setup of our experiments, i.e., the benchmarks and corresponding evaluation methodology and discuss experimental results. All experiments were performed on single NVIDIA Tesla (V100) GPU.  We also give an example of flatness analysis in the Appendix A.
\subsection{Text Classification}
\paragraph{Data.}We conduct experiments on 7 classification tasks from the GLUE benchmark~\citep{glue}. These datasets can be broadly divided into 3 families of problems. Single sentence tasks which include linguistic acceptability (CoLA) and sentiment analysis (SST-2), pairwise classification which include Similarity (MRPC) and paraphrasing (QQP) and inference tasks which include Natural Language Inference (MNLI and RTE) and Question Answering NLI (QNLI).Following prior works~\citep{bert, roberta}, we report  Matthews correlation for CoLA, F1 for QQP and MRPC and accuracy for the rest.

\paragraph{Experimental Setup.} We choose the pre-trained DistilRoBERTa (6-layer)~\citep{distilbert} as the base model for our experiments. We use a RoBERTa-Large model (24-layer)~\citep{roberta} fine-tuned on each of the tasks as the corresponding teacher model in the KD baseline. All models are trained with the AdamW optimizer with the default settings~\citep{adamw} for 30 epochs\footnote{Our experiments were conduct with huggingface toolkit (https://github.com/huggingface/transformers).}. To explore the best hyper-parameters, we used a grid search over the learning rate $\in \{1e-5, 2e-5, 3e-5\}$, batch size $\in \{8, 16, 32, 64\}$, KD interpolation ($\alpha) \in \{0.1, 0.2, 0.5\}$, KD temperature $\in \{1, 5, 10\}$ and the hyper-parameter $\rho \in \{0.05, 0.15\}$ for SAM.

   
We compare SWA with SAM~\cite{sam_lm} and three KD techniques: vanilla KD, Annealing KD (AKD)~\citep{annealing_kd}, and TAKD\citep{takd}.

\paragraph{Results}
Table~\ref{tab:glue_test} shows the GLUE leaderboard test results. We  observe that SWA outperforms fine-tuning on all datasets, which indicates its general applicability and, on average, SWA performance is comparable to SOTA KD methods.


\begin{table}[!thb]
\begin{center}\small
\caption{Performance  on SQuAD (v1 and v2) datasets.}
\label{tab:squad}
\begin{tabular}{lcccc}
\toprule
         & \multicolumn{2}{c}{\textbf{SQuAD}\_v1}       & \multicolumn{2}{c}{\textbf{SQuAD}\_v2}       \\\cmidrule(lr){2-3}\cmidrule(lr){4-5}
    \textbf{Model}          & EM        & F1    & EM        & F1   \\\midrule
DistilRoBERTa & 80.5     & 87.8  & 71.8     & 75.2 \\
+SWA          & \textbf{{81.1}}     & \textbf{{88.2}} & \textbf{{72.4}}     & \textbf{{75.7}} \\
\midrule
RoBERTa-base & 85.9	&92.0 
 & 79.8 &	83.2 \\
+SWA        & \textbf{{86.1}}    & \textbf{{92.2}}   & \textbf{{80.1}}     & \textbf{{83.4}} \\ 	
\bottomrule
\end{tabular}
\end{center}
\end{table}

\subsection{Question Answering}
We conduct experiments on two Question Answering benchmarks, SQuAD v1~\citep{squad_v1} and SQuAD v2~\citep{squad_v2} and report the exact match and F1-score on development sets. We choose the pre-trained DistilRoBERTa (6-layer)~\citep{distilbert} and RoBERTa ~\citep{roberta} as the backbone models for our QA experiments.

Table~\ref{tab:squad} shows the validation results on the two SQuAD datasets. We applied SWA to RoBERTa model with different sizes (6 and 12 layers). We observed that SWA can consistently outperform regular fine-tuning method on both versions of SQuAD and the results show that SWA provides larger improvement for a small model.  

\subsection{Summarization}
\begin{table}[!th]
\begin{center}\small
\caption{Performance  on XSum datasets.}
\label{tab:xsum}
\begin{tabular}{lccc}
\toprule
& \multicolumn{3}{c}{\textbf{ROUGE} }\\
\cmidrule(lr){2-4}
\textbf{Model}                    &  1 &  2 &  L \\ \midrule
T5-small                     & 31.33 &  9.26 &  24.19 \\
+ SWA & \textbf{32.01}               & \textbf{9.79}               & \textbf{24.75}              \\
\bottomrule
\end{tabular}
\end{center}
\end{table}
We conduct experiments on XSum~\citep{xsum} dataset with T5-small~\citep{t5} model. The models are evaluated based on ROUGE scores~\citep{rouge}. Specifically, we report ROUGE-1, ROUGE-2, and ROUGE-L, which compute
the uniform, bi-gram, and longest common sequence overlap with the reference summaries.

Table~\ref{tab:xsum} shows our results. We observe that
SWA improves the overall scores for T5 model over fine-tuning.

\section{Analysis}

\subsection{Constant LR vs Cyclical LR} 
\begin{figure}[!h]
    \centering
    \includegraphics[width=0.4\textwidth]{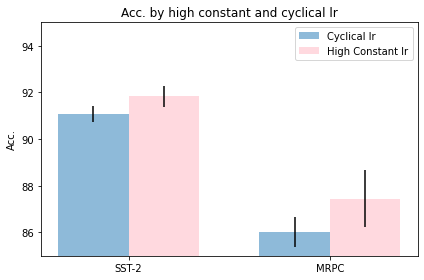}
    \caption{ Accuracy of DistilRoBERTa models for SST-2 and MRPC with cyclical and high constant learning rates. }
    \label{fig:ablation}
    
\end{figure}

We compare the effect of constant and cyclical learning rates on the SST-2 and MRPC datasets from GLUE on Figure~\ref{fig:ablation}. We can observe that a high constant learning rate performs much better than the cyclical one. We conducted an extensive grid search for $(\eta_{\text{max}}, \eta_{\text{min}}) \in \{(2e-5, 1e-6), (1e-5, 1e-6), (5e-6, 1e-6), (3e-6,1e-6)\}$ and $\eta_{c} \in \{2e-6, 3e-6\}$ and report the averaged accuracy and corresponding standard derivations of each method.  The patterns show the superior performance of high constant learning rate. 
{}
\subsection{Time Analysis of SWA.}

 \begin{table}[!th]
\caption{Time analysis for DistilRoBERTa.}
\label{tab: time}
\begin{center}\small
\begin{tabular}{l r}
\toprule
Method & relative time \\\midrule
Finetune & 1$\times$ \\
KD & 2.85$\times$ \\
AKD & 2.51$\times$ \\
SAM & 2.11$\times$ \\
SWA &1.08$\times$ \\
\bottomrule
\end{tabular}   
\end{center}
\end{table}

Table~\ref{tab: time} shows the relative time analysis for DistilRoBERTa model with different methods.
It is worth mentioning that, unlike KD methods, SWA doesn't need a trained teacher model, and, unlike SAM, it doesn't require double forward and backward pass computations. 
\section{Conclusion}

In this work we have adapted the SWA algorithm to the fine-tuning of PLMs. SWA is from a family of algorithms that aim to improve generalization for deep neural networks by finding a flatter, as opposed to a sharp, minimum. We present the recipe that can adapt SWA to fine-tuning PLMs, i.e. a high constant learning rate, averaging over training steps and choosing more components for discrimination and less for generation.

On compact PLMs, our adaptation of SWA performs on par with SOTA KD methods on the GLUE benchmark, and improves on fine-tuning on question answering and text summarization. Moreover, unlike KD or SAM (another flatness seeking optmization algorithm), SWA introduces no additional computation cost. As  future work, we want to explore apply to the pre-training of LMs.
\section*{Acknowledgments}
We thank Mindspore,\footnote{\url{https://www.mindspore.cn/}} which is a new deep learning computing framework, for partial support of this work. 
\section*{Limitations}
In the current work, we adapt a weigh assembling method, SWA, to PLMs fine-tuning and conduct extensive experiments to show that it can improve the generalization ability of PLMs. However, due to the limited computation resources, we don't explore the SWA for PLMs with a larger model size. 

\bibliography{anthology,custom}
\bibliographystyle{acl_natbib}

\appendix
\section{Flatness analysis}
To compare the flatness of fine-tuning and SWA, we compute the Hessian matrix $H$ for two minima found by fine-tuning and SWA on MRPC training dataset.
We consider two widely used measures of flatness: the largest eigenvalue of the Hessian matrix $\lambda_{\text{max}}(H)$ and the trace of Hessian. The results are shown in Table~\ref{tab: flatness}. It is clear that the minimum found by SWA has smaller $\lambda_{\text{max}}(H)$ and trace$(H)$, thus being flatter.
\begin{table}[!h]
\caption{Flatness of fine-tune and SWA on MRPC. To compute the Hessian, we exclude the embedding layers of DistilRoBERTa including word and position embedding, etc.}
\label{tab: flatness}
    \centering
    \begin{tabular}{l|c c}\toprule
       \textbf{Method}  & $\lambda_{\text{max}}(H)$ & trace$(H)$ \\ \midrule
       \textbf{Fine-tune}  & 60.22& 76.4 \\
       \textbf{SWA} & 1.19 & 4.72 \\ \bottomrule
    \end{tabular}
\end{table}



\end{document}